%% file: main.tex
\documentclass[10pt,twocolumn,letterpaper]{article}
\usepackage[pagenumbers]{cvpr} %

\makeatletter
\@namedef{ver@everyshi.sty}{}
\makeatother

\usepackage{graphicx}
\usepackage{amsmath}
\usepackage{amssymb}
\usepackage{booktabs}
\usepackage{xcolor}
\usepackage{overpic}
\usepackage{multirow,tabularx}
\usepackage{gensymb}
\usepackage{pifont}
\usepackage{adjustbox}
\usepackage{array}
\usepackage{nicematrix}
\usepackage[usestackEOL]{stackengine}
\usepackage{xpatch}
\usepackage[font=small,skip=5pt,belowskip=-8pt]{caption}
\usepackage[accsupp]{axessibility}

\usepackage[pagebackref,breaklinks,colorlinks]{hyperref}

\usepackage[capitalize]{cleveref}
\crefname{section}{Sec.}{Secs.}
\Crefname{section}{Section}{Sections}
\Crefname{table}{Table}{Tables}
\crefname{table}{Tab.}{Tabs.}

\def\cvprPaperID{2851} %
\def\confName{CVPR}
\def\confYear{2022}

\begin{document}
\input{definitions}

\title{Photorealistic Monocular 3D Reconstruction of Humans Wearing Clothing 
}

\author{Thiemo Alldieck \and Mihai Zanfir \and Cristian Sminchisescu}

\makeatletter
\let\@oldmaketitle\@maketitle%
\renewcommand{\@maketitle}{
    \@oldmaketitle%
    \centering
    \vspace{-7mm}
    {\bf Google Research}\\
    {\tt\small \{alldieck,mihaiz,sminchisescu\}@google.com}
    \vspace{10mm}
}
\makeatother

\maketitle

\begin{abstract}
    \vspace{-2mm}
    We present PHORHUM, a novel, end-to-end trainable, deep neural network methodology for photorealistic 3D human reconstruction given just a monocular RGB image.
    Our pixel-aligned method estimates detailed 3D geometry and, for the first time, the unshaded surface color together with the scene illumination. 
    Observing that 3D supervision alone is not sufficient for high fidelity color reconstruction, we introduce patch-based rendering losses that enable reliable color reconstruction on visible parts of the human, and detailed and plausible color estimation for the non-visible parts.
    Moreover, our method specifically addresses methodological and practical limitations of prior work in terms of representing geometry, albedo, and illumination effects, in an end-to-end model where factors can be effectively disentangled.
    In extensive experiments, we demonstrate the versatility and robustness of our approach. Our state-of-the-art results validate the method qualitatively and for different metrics, for both geometric and color reconstruction.
\end{abstract}

\vspace{-7mm}
\section{Introduction}
\label{sec:intro}

\input{sections/intro}

\section{Related Work}
\label{sec:related}
\input{sections/related}

\section{Method}
\label{sec:method}
\input{sections/method}

\section{Experiments}
\label{sec:experiments}
\input{sections/experiments}

\section{Discussion and Conclusions}
\label{sec:conclusion}

\input{sections/conclusion}

\appendix
\section*{Supplementary Material}

\input{sections/suppl.tex}

{\small
\bibliographystyle{ieee_fullname}

\input{main.bbl}
}

\end{document}

%% file: definitions.tex
\newcommand{\TA}[1]{{\textcolor{purple}{[\textbf{TA:} #1]}}}
\newcommand{\MZ}[1]{{\textcolor{red}{[\textbf{MZ:} #1]}}}
\newcommand{\CS}[1]{{\textcolor{blue}{[\textbf{CS:} #1]}}}

\newcommand{\cmark}{\ding{51}}%
\newcommand{\xmark}{\ding{55}}%
\newcolumntype{R}[2]{%
    >{\adjustbox{angle=#1,lap=\width-(#2)}\bgroup}%
    l%
    <{\egroup}%
}
\newcommand*\rot{\multicolumn{1}{R{35}{1em}}}

\makeatletter
\xpatchcmd{\paragraph}{3.25ex \@plus1ex \@minus.2ex}{3pt plus 1pt minus 1pt}{\typeout{success!}}{\typeout{failure!}}
\makeatother

\definecolor{limegreen}{HTML}{badc58}
\definecolor{myyellow}{HTML}{f6e58d}

\newcommand{\cbest}[1]{\cellcolor{limegreen}\textbf{#1}}
\newcommand{\csecond}[1]{\cellcolor{myyellow}#1}

\newcommand{\best}[1]{\colorbox{limegreen}{\textbf{#1}}}
\newcommand{\second}[1]{\colorbox{myyellow}{#1}}

\renewcommand{\vec}[1]{\boldsymbol{#1}}
\newcommand{\mat}[1]{\mathbf{#1}}
\newcommand{\set}[1]{\mathcal{#1}}

\newcommand{\loss}{\mathcal{L}}
\newcommand{\lossweight}{\lambda}

\newcommand{\geo}{\mathcal{S}}
\newcommand{\mesh}{\mathcal{M}}
\newcommand{\mlp}{f}
\newcommand{\backbone}{g}
\newcommand{\fullbackbone}{G}
\newcommand{\shadingnet}{s}
\newcommand{\weights}{\vec{\theta}}
\newcommand{\feature}{\vec{z}}
\newcommand{\point}{\vec{x}}
\newcommand{\pointintersect}{\hat{\point}}
\newcommand{\gtpointintersect}{\bar{\point}}
\newcommand{\normal}{\vec{n}}
\newcommand{\gtnormal}{\bar{\normal}}
\newcommand{\image}{\mat{I}}
\newcommand{\pixel}{\vec{p}}
\newcommand{\pixelset}{\set{I}}
\newcommand{\mask}{\mat{M}}
\newcommand{\albedoimg}{\mat{A}}
\newcommand{\front}{f}
\newcommand{\back}{b}
\newcommand{\encoding}{\gamma}
\newcommand{\dist}{d}
\newcommand{\albedo}{\vec{a}}
\newcommand{\gtalbedo}{\bar{\albedo}}
\newcommand{\cam}{\pi}
\newcommand{\signlabel}{l}
\newcommand{\patch}{p}
\newcommand{\shading}{\vec{s}}
\newcommand{\shaded}{\vec{c}}
\newcommand{\ray}{\vec{r}}
\newcommand{\illum}{\vec{l}}
\newcommand{\illumnet}{l}
\newcommand{\rays}{\set{R}}
\newcommand{\sigmoid}{\phi}
\newcommand{\sharpness}{k}
\newcommand{\origin}{\vec{o}}
\newcommand{\sign}{\sigma}

%% file: sections/intro.tex
\begin{figure}
    \centering
    \includegraphics[width=\linewidth]{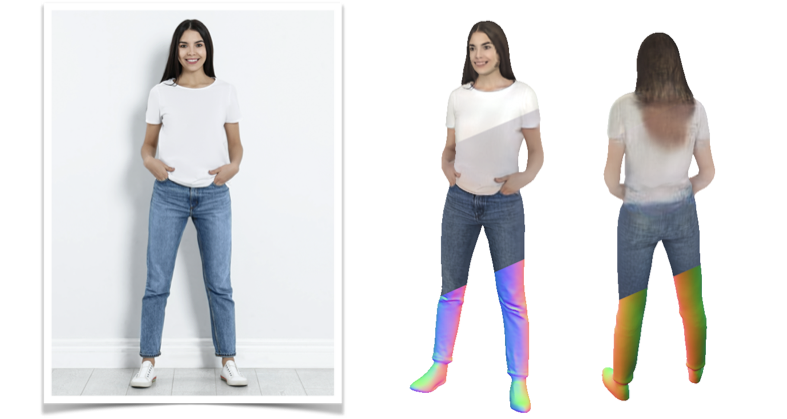}
    
    \caption{Given a single image, we reconstruct the full 3D geometry -- including self-occluded (or unseen) regions -- of the photographed person, together with albedo and shaded surface color. Our end-to-end trainable pipeline requires no image matting and reconstructs all outputs in a single step.}
    \label{fig:teaser}
\end{figure}

We present  PHORHUM, a method to photorealistically reconstruct the 3D geometry and appearance of a dressed person as photographed in a single RGB image.
The produced 3D scan of the subject not only accurately resembles the visible body parts but also includes plausible geometry and appearance of the non-visible parts, see fig.~\ref{fig:teaser}.
3D scans of people wearing clothing have many use cases and demand is currently rising.
Applications like immersive AR and VR, games, telepresence, virtual try-on, free-viewpoint photo-realistic
visualization, or creative image editing would all benefit from accurate 3D people models.
The classical way to obtain models of people is to automatically scan using multi-camera set-ups, manual creation by an artist, or a combination of both as often artists are employed to `clean up' scanning artifacts.
Such approaches are difficult to scale, hence we aim for alternative, automatic solutions that would be cheaper and easier to deploy.

Prior to us, many researchers have focused on the problem of human digitization from a single image \cite{varol2018bodynet,saito2019pifu,alldieck2019tex2shape,he2020geopifu,saito2020pifuhd,huang2020arch,he2021arch++}.
While these methods sometimes produce astonishingly good results, they have several shortcomings. First, the techniques often produce appearance estimates where shading effects are baked-in, and some methods do not produce color information at all.
This limits the usefulness of the resulting scans as they cannot be realistically placed into a virtual scene. Moreover, many methods rely on multi-step pipelines that first compute some intermediate representation, or perceptually refine the geometry using estimated normal maps.
While the former is at the same time impractical (since compute and memory requirements grow), and potentially sub-optimal (as often the entire system cannot be trained end-to-end to remove bias), the latter may not be useful for certain applications where the true geometry is needed, as in the case of body measurements for virtual try-on or fitness assessment, among others.
In most existing methods color is exclusively estimated as a secondary step.
However, from a methodological point of view, we argue that geometry and surface color should be computed simultaneously, since shading is a strong cue for surface geometry \cite{horn1970shape} and cannot be disentangled.

Our PHORHUM model specifically aims to address the above-mentioned state of the art shortcomings, as summarised in table~\ref{tab:features}. 
In contrast to prior work, we present an end-to-end solution that predicts geometry and appearance as a result of processing in a single composite network, with inter-dependent parameters, which are jointly estimated during a deep learning process.
The appearance is modeled as albedo surface color without scene specific illumination effects.
Furthermore, our system also estimates the scene illumination which makes it possible, in principle, to disentangle shading and surface color.
The predicted scene illumination can be used to re-shade the estimated scans, to realistically place another person in an existing scene, or to realistically composite them into a photograph.
Finally, we found that supervising the reconstruction using only sparse 3D information leads to perceptually unsatisfactory results.
To this end, we introduce rendering losses that increase the perceptual quality of the predicted appearance. %
Our contributions can be summarised as follows:
\begin{itemize}
    \itemsep-0.5mm
    \item[-] We present an end-to-end trainable system for high quality human digitization
    \item[-] Our method computes, for the first time, albedo and shading information
    \item[-] Our rendering losses significantly improve the visual fidelity of the results
    \item[-] Our results are more accurate and feature more detail than current state-of-the-art
\end{itemize}
\vspace{-2mm}

\begin{table}
    \begin{center}
    \small
    \begin{tabular}{ccccccc|l}
    \rot{end-to-end trainable} & \rot{signed distances}  & \rot{returns color} & \rot{returns albedo}  & \rot{returns shading} & \rot{true surface normals} & \rot{no mask needed} & \\
    \hline\hline
    \xmark & \xmark & \cmark & \xmark & \xmark & \cmark & \xmark & PIFu \cite{saito2019pifu} \\
    \xmark & \xmark & \xmark & \xmark & \xmark & \cmark & \cmark & PIFuHD \cite{saito2020pifuhd} \\
    \xmark & \xmark & \xmark & \xmark & \xmark & \cmark & \xmark & Geo-PIFu \cite{he2020geopifu} \\
    \xmark & \xmark & \cmark & \xmark & \xmark & \xmark  & \xmark & Arch \cite{huang2020arch} \\
    \xmark & \xmark & \cmark & \xmark & \xmark & \xmark & \xmark & Arch++ \cite{he2021arch++} \\
    \hline
    \cmark & \cmark & \cmark & \cmark & \cmark & \cmark & \cmark & \textbf{PHORHUM (ours)} \\
    \end{tabular}
    \end{center}
    \vspace{-3mm}
    \caption{Overview of the properties of single image 3D human reconstruction methods. Our method is the only one predicting albedo surface color and shading. Further, our method has the most practical training set-up, does not require image matting at test-time, and returns signed distances rather than binary occupancy -- a more informative representation.}
    \label{tab:features}
\end{table}

%% file: sections/related.tex
Reconstructing the 3D shape of a human from a single image or a monocular video is a wide field of research. Often 3D shape is a byproduct of 3D human pose reconstruction and is represented trough parameters of a statistical human body model \cite{smpl2015loper,xu2020ghum}. In this review, we focus on methods that go beyond and reconstruct the 3D human shape as well as garments or hairstyle.
Early pioneering work is optimization-based. Those methods use videos of moving subjects and integrate information over time in order to reconstruct the complete 3D shape \cite{Bogo:ICCV:2015, alldieck2018video}.
The advent of deep learning questioned the need for video.
First, hybrid reconstruction methods based on a small number of images have been presented \cite{alldieck2019learning, bhatnagar2019mgn}.
Shortly after, approaches emerged to predict 3D human geometry from a single image.
Those methods can be categorized by the used shape representation: 
voxel-based techniques \cite{Zheng2019DeepHuman,varol2018bodynet,jackson20183d} predict whether a given segment in space is occupied by the 3D shape.
A common limitation is the high memory requirement resulting in shape estimates of limited spatial resolution.
To this end, researchers quickly adopted alternative representations including visual hulls \cite{natsume2019siclope}, moulded front and back depth maps \cite{gabeur2019moulding, smith2019facsimile}, or augmented template meshes \cite{alldieck2019tex2shape}.
Another class of popular representations consists of implicit function networks (IFNs).
IFNs are functions over points in space and return either whether a point is inside or outside the predicted shape \cite{i_OccNet19, i_IMGAN19} or return its distance to the closest surface \cite{i_DeepSDF}.
Recently IFNs have been used for various 3D human reconstruction tasks \cite{deng2019nasa,chibane20ifnet,gropp2020igr,mustafa2021multi} and to build implicit statistical human body models \cite{LEAP:CVPR:2021,alldieck2020imghum}.
Neural radiance fields \cite{mildenhall2020nerf} are a related class of representations specialized for image synthesis that have also been used to model humans \cite{peng2021neural,xu2021hnerf,liu2021neural}.
Saito \etal were the first to use IFNs for monocular 3D human reconstruction.
They proposed an implicit function conditioned on pixel-aligned features \cite{saito2019pifu, saito2020pifuhd}.
Other researchers quickly adopted this methodology for various use-cases \cite{li2020monocular, he2020geopifu, zheng2021pamir, yang2021s3}. ARCH \cite{huang2020arch} and ARCH++ \cite{he2021arch++} also use pixel-aligned features but transform information into a canonical space of a statistical body model.
This process results in animatable reconstructions, which comes, however, at the cost of artifacts that we will show. In this work, we also employ pixel-aligned features but go beyond the mentioned methods in terms of reconstructed surface properties (albedo and shading) and in terms of the quality of the 3D geometry.
Also related is H3D-Net \cite{ramon2021h3d}, a method for 3D head reconstruction, which uses similar rendering losses as we do, but requires three images and test-time optimization. In contrast, we work with a monocular image, purely feed-forward.

%% file: sections/method.tex
\begin{figure*}
    \centering
    \begin{overpic}[width=1.\linewidth]{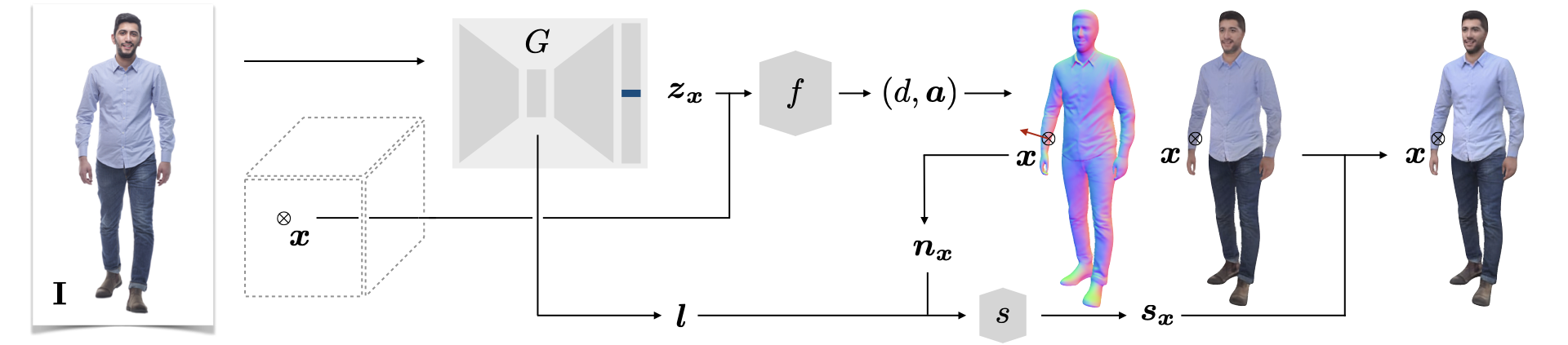} %
    \end{overpic}%
    \caption{Overview of our method. The feature extractor network $\fullbackbone$ produces pixel-aligned features $\feature_{\point}$ from an input image $\image$ for all points in space $\point$. The implicit signed distance function network $\mlp$ computes the distance $\dist$ to the closest surface given a point and its feature. Additionally $\mlp$ returns albedo colors $\albedo$ defined for surface points. The shading network $\shadingnet$ predicts the shading for surface points given its surface normal $\normal_{\point}$, as well as illumination $\illum$. On the right we show the reconstruction of geometry and albedo colors, and the shaded 3D geometry.}
    \label{fig:method}
\end{figure*}

Our goal is to estimate the 3D geometry $\geo$  of a subject as observed in a single image $\image$. Further, we estimate the unshaded albedo surface color and a per-image lighting model.
$\geo$ is defined as the zero-level-set of the signed distance function (SDF) $\mlp$ represented using a neural network,
\begin{equation}
    \geo_{\weights}(\image) = \Big\{ \point \in \mathbb{R}^3 \: | \: \mlp\big( \backbone(\image, \point; \weights), \encoding(\point); \weights \big) = (0, \albedo) \Big\}
\end{equation}
where $\weights$ is the superset of all learnable parameters. 
The surface $\geo$ is parameterized by pixel aligned features $\feature$ (\cf \cite{saito2019pifu}) computed from the input image $\image$ using the feature extractor network $\fullbackbone$
\begin{equation}
     \backbone(\image, \point; \weights) = b(\fullbackbone(\image; \weights), {\cam(\point)}) = \feature_{\point},
\end{equation}
where $b$ defines pixel access with bilinear interpolation and $\cam(\point)$ defines the pixel location of the point $\point$ projected using camera $\cam$.
$\mlp$ returns the signed distance $\dist$ of the point $\point$ \wrt $\geo$ and additionally its albedo color $\albedo$
\begin{equation}
     \mlp(\feature_{\point}, \encoding(\point); \weights) = (\dist, \albedo),
\end{equation}
where $\encoding$ denotes basic positional encoding as defined in \cite{tancik2020fourfeat}. In the sequel, we will use $\dist_{\point}$ for the estimated distance at $\point$ and $\albedo_{\point}$ for the color component, respectively.

To teach the model to decouple shading and surface color, we additionally estimate the surface shading using a per-point surface shading network
\begin{equation}
     \shadingnet(\normal_{\point}, \illum; \weights) = \shading_{\point},
\end{equation}
where $\normal_{\point} = \nabla_{\point} \dist_{\point}$ is the estimated surface normal defined by the gradient of the estimated distance \wrt $\point$.
$\illumnet(\image; \weights) = \illum$ is the illumination model estimated from the image. In practice, we use the bottleneck of $\fullbackbone$ for $\illum$ and further reduce its dimensionality.
The final shaded color is then $\shaded = \shading \circ \albedo$ with $\circ$ denoting element-wise multiplication. We now define the losses we use to train $\mlp$, $\fullbackbone$, and $\shadingnet$.

\subsection{Losses}
\label{sec:losses}
We create training examples by rendering scans of humans and drawing samples from the raw meshes -- please see \S\ref{sec:dataset} for details. We define losses based on sparse 3D supervision and losses informed by ray-traced image patches.

\noindent\textbf{Geometry and Color Losses.}
Given a ground truth mesh $\mesh$ describing the surface $\geo$ as observed in an image $\image$ and weights $\lossweight_{*}$ we define losses as follows. The surface is supervised via samples $\set{O}$ taken from the mesh surface $\mesh$ and enforcing their distance to return zero and the distance gradient to follow their corresponding ground truth surface normal $\gtnormal$
\begin{equation}
    \vspace{-1mm}
    \loss_{g} = \frac{1}{|\set{O}|} \sum_{i \in \set{O}} \lossweight_{g_1}  |\dist_{\point_i}| + \lossweight_{g_2} \| \normal_{\point_i} - \gtnormal_i \|.
\end{equation}
Moreover, we supervise the sign of additional samples $\set{F}$ taken around the surface
\begin{equation}
    \loss_{l} = \frac{1}{|\set{F}|} \sum_{i \in \set{F}} \text{BCE}\big(\signlabel_i, \sigmoid(\sharpness \dist_{\point_i})\big),
\end{equation}
where $\signlabel$ are inside/outside labels, $\sigmoid$ is the sigmoid function, and $\text{BCE}$ is the binary cross-entropy. $\sharpness$ determines the sharpness of the decision boundary and is learnable. %
Following \cite{gropp2020igr}, we apply geometric regularization such that $\mlp$ approximates a SDF with gradient norm $1$ everywhere
\begin{equation}
    \loss_{e} = \frac{1}{|\set{F}|} \sum_{i \in \set{F}} (\| \normal_{\point_i} \| - 1)^2.
\end{equation}
Finally, we supervise the albedo color with the `ground truth' albedo $\gtalbedo$ calculated from the mesh texture
\begin{equation}
    \loss_{a} = \lossweight_{a_1} \frac{1}{|\set{O}|} \sum_{i \in \set{O}} | \albedo_{\point_i} - \gtalbedo_i | + \lossweight_{a_2} \frac{1}{|\set{F}|} \sum_{i \in \set{F}} | \albedo_{\point_i} - \gtalbedo_i |.
\end{equation}
Following \cite{saito2019pifu}, we apply $\loss_{a}$ not only on but also near the surface. Since albedo is only defined on the surface, we approximate the albedo for points near the surface with the albedo of their nearest neighbor on the surface.

\noindent\textbf{Rendering losses.}
The defined losses are sufficient to train our networks. However, as we show in the sequel, 2D rendering losses help further constrain the problem and increase the visual fidelity of the results.
To this end, during training, we render random image patches of the surface $\geo$ with random strides and fixed size using ray-tracing.
First, we compute the rays $\rays$ corresponding to a patch as defined by $\cam$. We then trace the surface using two strategies.
First, to determine if we can locate a surface along a ray, we query $\mlp$ in equal distances along every ray $\ray$ and compute the sign of the minimum distance value
\begin{equation}
    \sign_{\ray} = \sigmoid \big(\sharpness \min_{t\ge 0} \dist_{\origin + t \ray}\big),
\end{equation}
where $\origin$ is the camera location. We then take the subset $\rays_{\geo} \subset \rays$ of the rays containing rays where $\sign \le 0.5$ and $\signlabel = 0$, \ie we select the rays which located a surface where a surface is expected.
Hereby, the inside/outside labels $\signlabel$ are computed from pixel values of the image segmentation mask $\mask$ corresponding to the rays. For the subset $\rays_{\geo}$, we exactly locate the surface using sphere tracing. Following \cite{yariv2020multiview}, we make the intersection point $\pointintersect$ at iteration $t$ differentiable \wrt to the network parameters without having to store the gradients of sphere tracing
\begin{equation}
    \pointintersect = \pointintersect^t - \frac{\ray}{\normal^t \cdot \ray} \dist_{\pointintersect^t}.
\end{equation}
In practice, we trace the surface both from the camera into the scene and from infinity back to the camera. This means, we locate both the front surface and the back surface. We denote the intersection points $\pointintersect^{\front}$ for the front side and $\pointintersect^{\back}$ for the back side, respectively.
Using the above defined ray set $\rays_{\geo}$ and intersection points $\pointintersect$, we enforce correct surface colors through
\begin{equation}
    \loss_{r} = \frac{1}{|\rays_{\geo}|} \sum_{i \in \rays_{\geo}} | \albedo_{\pointintersect_i^{\front}} - \gtalbedo_i^\front| + | \albedo_{\pointintersect_i^{\back}} - \gtalbedo_i^\back|,
\end{equation}
where ground truth albedo colors $\gtalbedo$ are taken from synthesized unshaded images $\albedoimg^\front$ and $\albedoimg^\back$. The back image $\albedoimg^\back$ depicts the backside of the subject and is created by inverting the Z-buffer during rendering. We explain this process in more detail in \S\ref{sec:dataset}. Additionally, we also define a VGG-loss \cite{chen2017photographic} $\loss_{\text{VGG}}$ over the rendered front and back surface patches, enforcing that structure is similar to the unshaded ground-truth images.
Finally, we supervise the shading using
\vspace{-2mm}
\begin{equation}
    \loss_{c} = \frac{1}{|\rays_{\geo}|} \sum_{i \in \rays_{\geo}} | \albedo_{\pointintersect_i^{\front}} \circ \shading_{\pointintersect_i} - \pixel_i|,
\end{equation}
with $\pixel$ being the pixel color in the image $\image$ corresponding to the ray $\ray$. We found it also useful to supervise the shading on all pixels of the image $\pixelset = \{\pixel_0, \dots, \pixel_N\}$ using ground truth normals $\gtnormal$ and albedo $\gtalbedo$
\begin{equation}
    \loss_{s} = \frac{1}{N} \sum_{i \in \pixelset} | \gtalbedo_i^\front \circ \shadingnet(\gtnormal_i, \illum; \weights) - \pixel_i|.
\end{equation}
The final loss is a weighted combination of all previously defined losses $\loss_{*}$. In \S\ref{sec:ablations}, we ablate the usage of the rendering losses and the shading estimation network.

 \begin{figure}
    \centering
    \includegraphics[width=\linewidth]{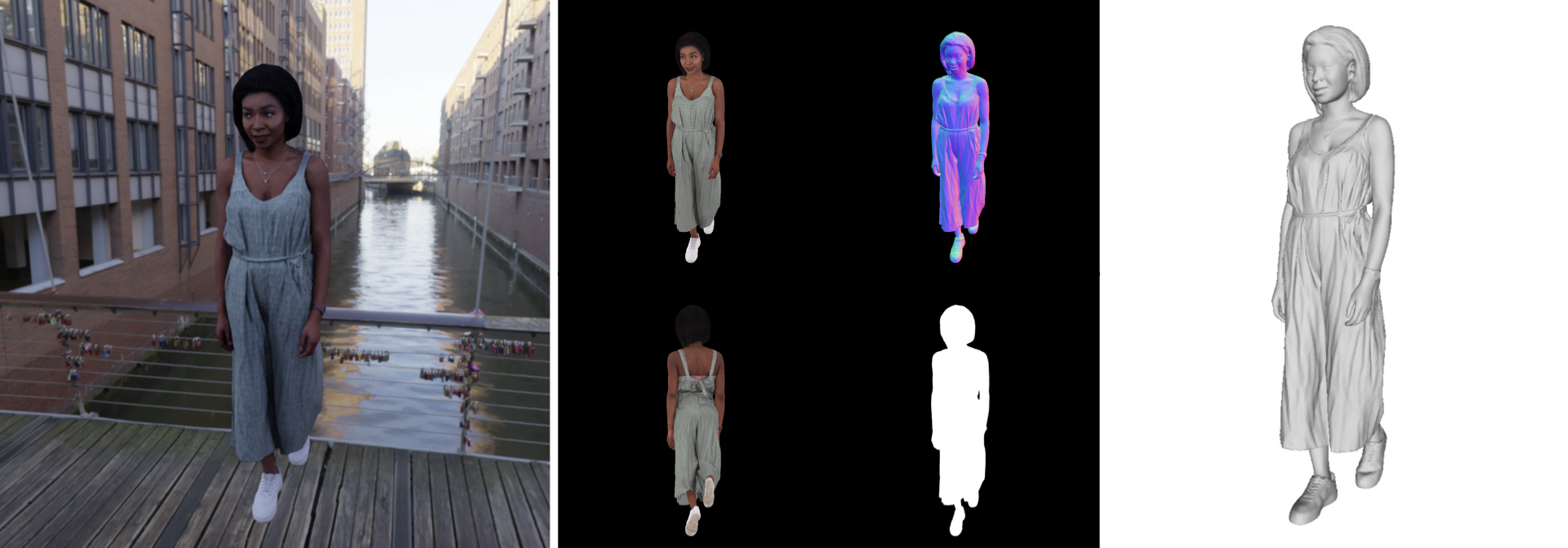}
    
    \caption{A sample from our dataset. From left to right: rendered, shaded image on HDRI background; front and back albedo images; normal and an alpha map, and 3D mesh used for sampling.}
    \label{fig:datasample}
\end{figure}

\subsection{Dataset}
\label{sec:dataset}
We train our networks using pairs of meshes and rendered images.
The meshes are scans of real people from commercial websites \cite{renderpeople} and our own captured data. %
We employ high dynamic range images (HDRI) \cite{hdri} for realistic image-based lighting and as backgrounds.
Additionally to the shaded images, we also produce an alpha mask and unshaded albedo images.
In the absence of the true surface albedo, we use the textures from the scans.
Those are uniformly lit but may contain small and local shading effects, \eg from small wrinkles.
As mentioned earlier, we produce not only a front side albedo image, but also one showing the back side. 
We obtain this image by inverting the Z-buffer during rendering. This means, not that the first visible point along each camera ray is visible, but the last passed surface point. See fig.~\ref{fig:datasample} for an example of our training images.
Furthermore, we produce normal maps used for evaluation and to supervise shading.
Finally, we take samples by computing 3D points on and near the mesh surface and additionally sample uniformly in the bounding box of the whole dataset.
For on-surface samples, we compute their corresponding albedo colors and surface normals, and for near and uniform samples we compute inside/outside labels by casting randomized rays and checking for parity.

We use $217$ scans of people in different standing poses, wearing various outfits, and sometimes carrying bags or holding small objects. The scans sources allow for different augmentations: we augment the outfit colors for 100 scans and repose 38 scans. In total we produce a dataset containing $\approx190$K images, where each image depicts a scan rendered with a randomly selected HDRI backdrop and with randomized scan placement.
Across the $217$ scans some share the same identity. We strictly split test and train identities and create a test-set containing 20 subjects, each rendered under 5 different light conditions.

\subsection{Implementation Details}
\label{sec:imp_details}
We now present our implementation and training procedure.
Our networks are trained with images of $512\times512$px resolution.
During training we render $32\times32$px patches with stride ranging from zero to three. We discard patches that only include background.
Per training example we draw random samples for supervision from the surface and the space region around it. Concretely, we draw each 512 samples from the surface, near the surface and uniformly distributed over the surrounding space.
The samples are projected onto the feature map using a projective camera with fixed focal length.

The feature extractor $\fullbackbone$ is a U-Net with 13 encoder-decoder layers and skip connections.
The first layer contains 64 filters and the filter size is doubled in the encoder in each layer up to 512 at the maximum.
The decoder halves the filter size at the 11th layer, which effectively means that $\fullbackbone$ produces features in $\mathbb{R}^{256}$.
We use Leaky ReLU activations and blur-pooling \cite{zhang2019making} for the encoder and bilinear resizing for the decoder, respectively.
The geometry network $\mlp$ is a MLP with eight 512-dimensional fully-connected layers with Swish activation \cite{ramachandran2017searching}, an output layer with Sigmoid activation for the color component, and a skip connection to the middle layer. %
The shading network $\shadingnet$ is conditioned on a 16-dimensional illumination code and consists of three 256-dimensional fully-connected layers with Swish activation and an output layer with ReLU activation.
Our total pipeline is relatively small and has only 48.8M trainable parameters.
We train all network components jointly, end-to-end, for 500k iterations using the Adam optimizer \cite{kingma2014adam}, with learning-rate of $1\times10^{-4}$, linearly decaying with a factor of $0.9$ over 50k steps.
Please refer to our supplementary material for a list of our loss weights $\lossweight_*$.

%% file: sections/experiments.tex
We present quantitative evaluation results and ablation studies for geometric and color reconstruction on our own dataset. We also show qualitative results for real images. 

\paragraph{Inference.} At inference time, we take as input an RGB image of a person in a scene. Note that we do not require the foreground-background mask of the person. However, in practice we use a bounding box person detector to center the person and crop the image -- a step that can also be performed manually. We use Marching Cubes \cite{lewiner2003efficient} to generate our reconstructions by querying points in a 3D bounding box at a maximum resolution of $512^3$. We first approximate the bounding box of the surface by probing at coarse resolution and use Octree sampling to progressively increase the resolution as we get closer to the surface. This allows for very detailed reconstructions of the surface geometry with a small computational overhead, being made possible by the use of signed distance functions in our formulation.

\paragraph{Camera Model.} Different from other methods in the literature, we deviate from the standard orthographic camera model and instead use perspective projection, due to its general validity.
A model assuming an orthographic camera would in practice produce incorrect 3D geometry. In fig.~\ref{fig:qualitiative_comp_pifuhd} one can see the common types of errors for such models. The reconstructed heads are unnaturally large, as they extend in depth away from the camera. In contrast, our reconstructions are more natural, with correct proportions between the head and the rest of the body.

\paragraph{Competing Methods.} We compare against other single-view 3D reconstructions methods that leverage pixel-aligned image features. PIFu~\cite{saito2019pifu} is the pioneering work and learns an occupancy field. PIFuHD~\cite{saito2020pifuhd}, a very parameter-heavy model, builds upon PIFu with higher resolution inputs and leverages a multi-level architecture for coarse and fine grained reconstruction. It also uses offline estimated front and back normal maps as additional input. GeoPIFu~\cite{he2020geopifu} is also a multi-level architecture, but utilizes latent voxel features as a coarse human shape proxy. ARCH~\cite{huang2020arch} and ARCH++~\cite{he2021arch++} transform information into the canonical space of a statistical body model. This sacrifices some of the reconstruction quality for the ability to produce animation-ready avatars. For PIFu, ARCH, ARCH++, an off-the-shelf detector \cite{kirillov2020pointrend} is used to segment the person in the image, whereas PHORHUM (us) and PIFuHD use the raw image. The results of ARCH and ARCH++ have been kindly provided by the authors.

Due to the lack of a standard dataset and the non-availability of training scripts of most methods, all methods have been trained with similar but different datasets. All datasets are sufficiently large to enable generalization across various outfits, body shapes, and poses.
Please note that our dataset is by far the smallest with only 217 scans. All other methods use $>400$ scans.

\begin{table}
    \begin{center}
    \small
    \centering
    \begin{NiceTabular}{ccc|l}[colortbl-like]
        Front side & Back side & Mean & \\
        \hline
        2.68 & 2.15 & 2.42 & PIFu \cite{saito2019pifu} \\
        2.51 & 2.04 & 2.28 & ARCH \cite{huang2020arch} \\
        2.68 & 2.26 & 2.47 & ARCH++ \cite{he2021arch++} \\
        \hline
        \textbf{2.88} & \textbf{2.43} & \textbf{2.65} & \textbf{PHORHUM (Ours)} \\
    \end{NiceTabular}
    \end{center}%
    \vspace{-3mm}
    \caption{Inception Score of renderings of the front and back side of the 3D reconstructions. Our method produces the most natural surface colors for both the front and the unseen back.}
    \label{tab:numerical_eval_is}
    \vspace{-1mm}
\end{table}

\subsection{Reconstruction Accuracy}

\begin{table*}
    \begin{center}
    \footnotesize
    \centering
    \def\arraystretch{1.1}%
    \setlength{\tabcolsep}{5pt}
    \begin{NiceTabular}{ccc|ccc|ccc|ccc|l}[colortbl-like]
    \Block{1-3}{3D Metrics} & & & \Block{1-3}{Rendered Normals} & & & \Block{1-3}{Shaded Rendering} & & & \Block{1-3}{Albedo Rendering} & & & \\
    Ch. $\downarrow$ & IoU $\uparrow$ & NC $\uparrow$ & SSIM $\uparrow$ & LPIPS $\downarrow$ & PSNR $\uparrow$ & SSIM $\uparrow$ & LPIPS $\downarrow$ & PSNR $\uparrow$ & SSIM $\uparrow$ & LPIPS $\downarrow$ & PSNR $\uparrow$ & \\
    \hline
    3.21 & 0.61 & 0.77 & 0.71 & 0.17 & 17.69 & \csecond{0.83} & \csecond{0.16} & \cbest{24.57} & -- & -- & -- & PIFu
    \cite{saito2019pifu} \\
    4.54 & \csecond{0.62} & 0.78 & \cbest{0.78} & \cbest{0.10} & \cbest{20.15}  & -- & -- & -- & -- & -- & -- & PIFuHD
    \cite{saito2020pifuhd} \\
    4.98 & 0.54 & 0.72 & 0.68 & 0.18 & 17.25 & -- & -- & -- & -- & -- & -- & Geo-PIFu
    \cite{he2020geopifu} \\
    3.58 & 0.57 & 0.75 & 0.68 & 0.20 & 15.51 & 0.72 & 0.23 & 19.28 & -- & -- & -- & ARCH
    \cite{huang2020arch} \\
    3.48 & 0.59 & 0.77 & 0.70 & 0.17 & 16.24 & \csecond{0.83} & 0.17 & 22.69 & -- & -- & -- &  ARCH++
    \cite{he2021arch++} \\
    \hline

    \cbest{1.14} & \cbest{0.73} & \csecond{0.84} &  %
    \csecond{0.77} & 0.12 & 19.25 & %
    -- & -- & -- & %
    \csecond{0.82} & 0.16 & 20.51 & Ours w/o rendering \\ %
    
    \csecond{1.29} & \cbest{0.73} & \cbest{0.85} & %
    \cbest{0.78} & \csecond{0.11} & \csecond{19.67} %
    & -- & -- & -- & %
    \cbest{0.85} & \csecond{0.13} &  \csecond{22.02} & Ours w/o shading \\ %
    
    \hline
    \csecond{1.29} & \cbest{0.73} & \cbest{0.85} & %
    \cbest{0.78} & \csecond{0.11} & 19.60 & %
    \cbest{0.85} & \cbest{0.13} & \csecond{24.01} & %
    \cbest{0.85} & \cbest{0.12} & \cbest{22.23} &  \textbf{PHORHUM (Ours)} \\ %
    \end{NiceTabular}
    \end{center}%
    \vspace{-3mm}
    \caption{Numerical comparisons with other single-view 3D reconstructions methods and ablations of our method. We mark the \best{best} and \second{second best} results. All Chamfer metrics are $\times10^{-3}$. }
    \label{tab:numerical_eval}
    \vspace{-1mm}
\end{table*}

To evaluate the geometric reconstruction quality, we report several metrics, namely: bi-directional Chamfer distance (Ch.~$\downarrow$), Normal Consistency (NC $\uparrow$), and Volumetric Intersection over Union (IoU $\uparrow$). To account for the inherent ambiguity of monocular reconstruction \wrt scale, we first use Iterative Closest Point to align the reconstructions with the ground truth shapes.
Additionally, we evaluate how well the visible part of the person is reconstructed. This also mitigates effects caused by camera model assumptions. We render the reconstruction under the assumed camera model and compare with the original image, the unshaded albedo image, and the rendered normals. For image reconstruction metrics, we use peak signal-to-noise ratio (PSNR $\uparrow$), structural similarity index (SSIM $\uparrow$) and learned perceptual image patch similarity (LPIPS $\downarrow$). Finally, we use the Inception Score (IS $\uparrow$) \cite{salimans2016improved} as a perceptual metric. This allows us to also evaluate non-visible parts where no ground truth is available, as in the case of the shaded backside view of a person.

We report the introduced metrics in tables \ref{tab:numerical_eval_is} and \ref{tab:numerical_eval}. Our model produces the most natural surface colors for both the visible front side and the non-visible back side.
Furthermore, our method produces the most accurate 3D reconstructions and is the only one that computes the surface albedo. Our results are on-par with those of PIFuHD in terms of surface normal reconstruction. In contrast to our method, PIFuHD specifically targets surface normals with a dedicated image-translation network. ARCH and ARCH++ also specifically handle surface normals, but in contrast to all other methods, only compute a normal map and do not refine the true geometry. Note that we use normal mapping (not true surface normals) for ARCH and ARCH++ in the comparison and in all the following figures.
For shaded rendering of the front side, the original PIFu is numerically on par with our method. However, the results are blurry, which is evident in the lower Inception Score and LPIPS. PIFu and all other competing methods do not decompose albedo and shading, which means that they can simply project the original image onto the reconstruction. Although our method performs a harder task, our results are among the best, or the best, across all metrics.

\subsection{Qualitative Results}

\begin{figure*}
    \centering
    \begin{overpic}[width=1.0\linewidth]{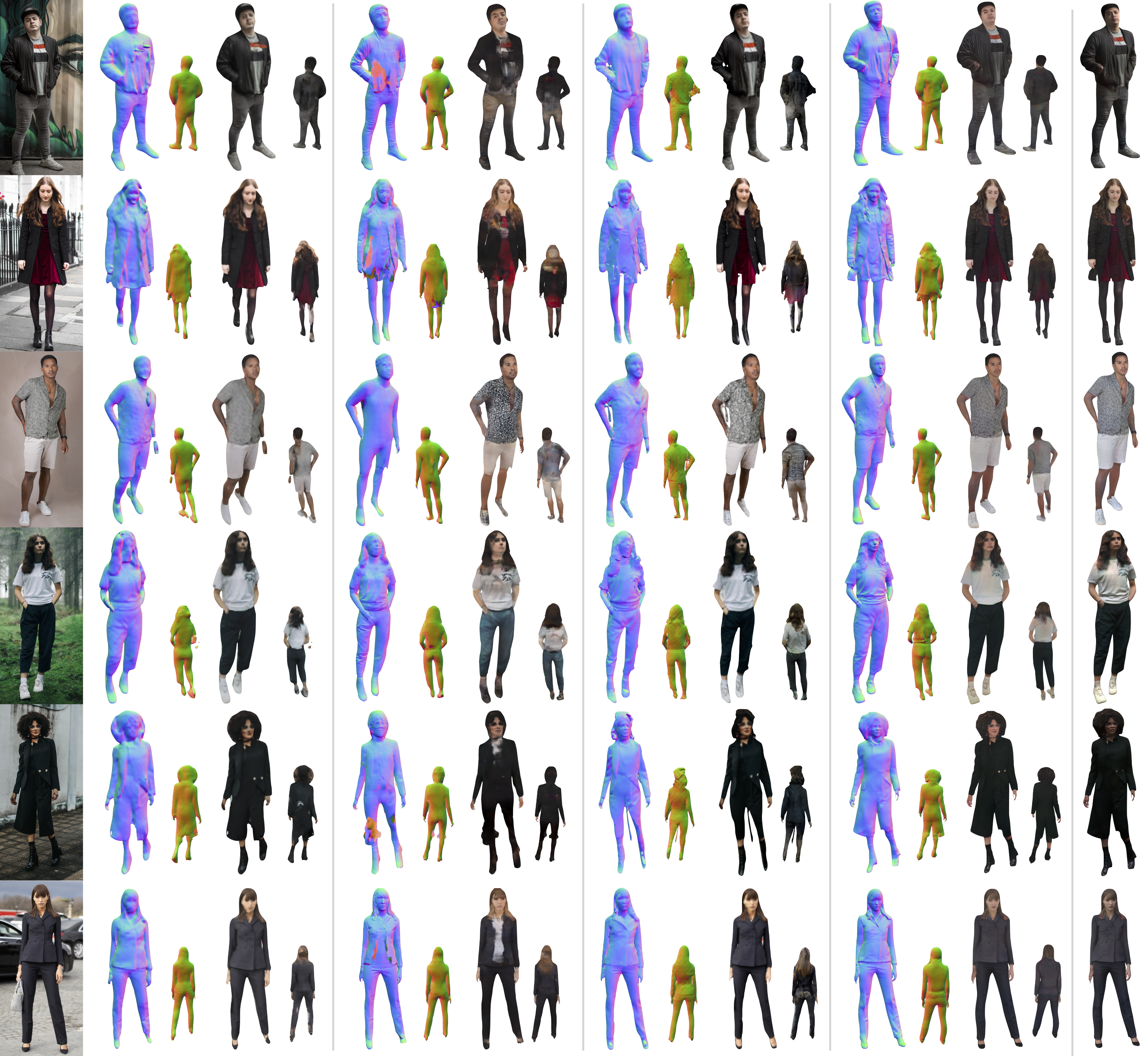}%
    \put(18,-1.5){\makebox[0pt]{\Centerstack{\footnotesize{PIFu}}}}
    \put(40,-1.5){\makebox[0pt]{\Centerstack{\footnotesize{ARCH}}}}
    \put(61.5,-1.5){\makebox[0pt]{\Centerstack{\footnotesize{ARCH++}}}}
    \put(83,-1.5){\makebox[0pt]{\Centerstack{\footnotesize{PHORHUM}}}}
    \end{overpic}%
    \vspace{2.5mm}
    \caption{Qualitative comparisons on real images with state-of-the-art methods that produce color. From left to right: Input image, PIFu, ARCH, ARCH++, PHORHUM (ours), our shaded reconstruction. For each method we show the 3D geometry and the reconstructed color. Our method produces by far the highest level of detail and the most realistic color estimate for the unseen back side.}
    \label{fig:qualitiative_comp}
\end{figure*}

\begin{figure*}
    \centering
    \includegraphics[width=\linewidth]{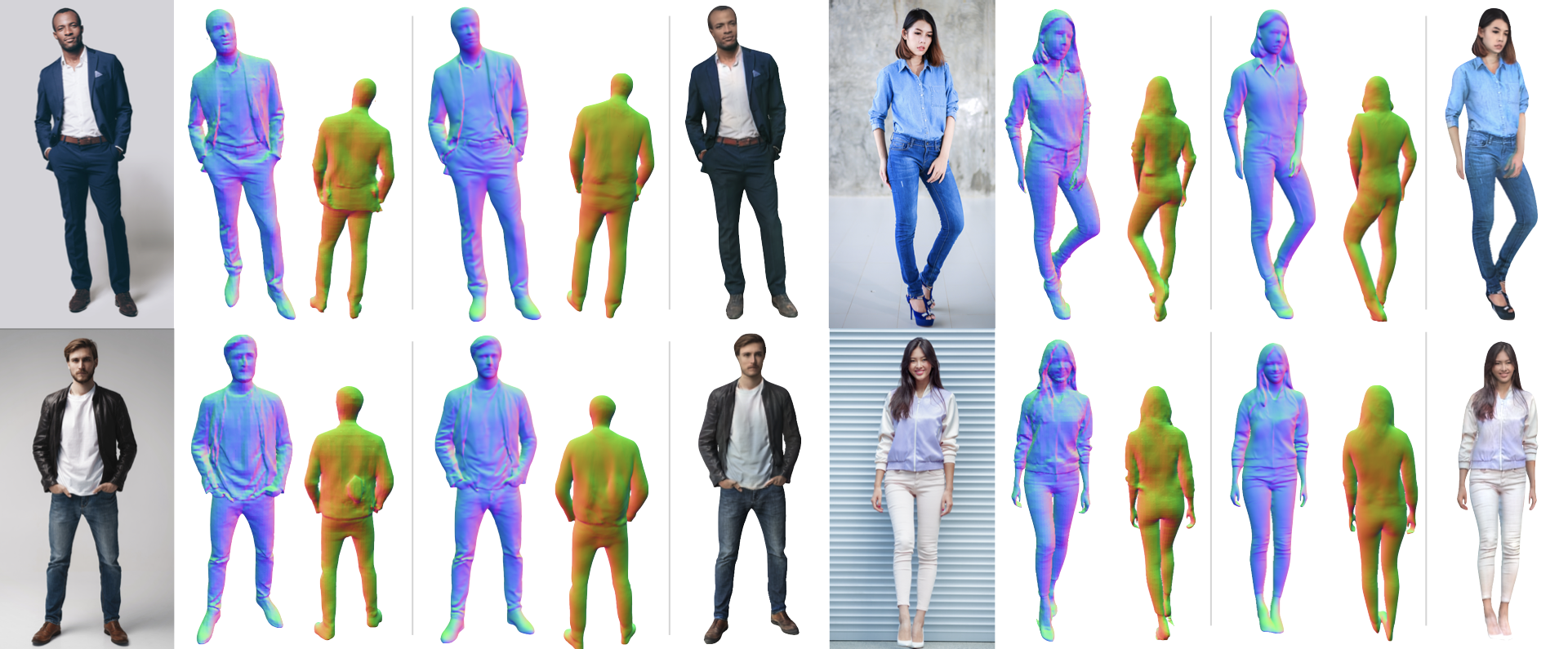}
    \caption{Qualitative comparisons on real images with the state-of-the-art method PIFuHD. We show front and back geometry produced by PIFuHD (left) and our results (right). Our reconstructions feature a similar level of detail but contain less noise and body poses are reconstructed more reliably. Additionally, our method is able to produce albedo and shaded surface color -- we show our shaded reconstructions for reference.}
    \label{fig:qualitiative_comp_pifuhd}
    \vspace{-1mm}
\end{figure*}

Quantitative evaluations do not always correlate well with human perception. To this end, we show qualitative results of our method and results of PIFu, ARCH, and ARCH++ on real images in fig.~\ref{fig:qualitiative_comp}, and a side-by-side comparison with PIFuHD in fig.~\ref{fig:qualitiative_comp_pifuhd}.

In fig.~\ref{fig:qualitiative_comp}, we show the 3D reconstructions with color-mapped normals, and the colored reconstructions, both front and back. For our method we render the albedo and additionally show the shaded reconstruction in the last column.
Our method reliably reconstructs facial detail, hair, and clothing wrinkles. The albedo features small color patterns visible in the input image and, at the same time, does not contain strong shading effects.
The reconstructed non-visible back side is sharp, detailed, and matches our expectations well.
The clothing items are well separated and small details like hair curls are present.
ARCH and ARCH++ encounter problems reconstructing the red dress in line two, sometimes produce artifacts, and fail entirely for the subject in line five.
The observed problems are common for methods that reconstruct relative to, or in the canonical space, of a body model.
In contrast, our method produces complete, smooth, and detailed reconstructions.

PIFuHD does not compute surface color, thus we only compare the geometry in fig.~\ref{fig:qualitiative_comp_pifuhd}. We show our shaded results only for completeness.
Consistent with the numerical results, our results are on par in terms of level of detail.
However, our reconstructions are smoother and contain less noise -- a property of signed distance functions. Our model is capable of producing these results by using a rather small network capacity.
In contrast PIFuHD is an extremely large model that is specifically tailored for surface normal estimation.

As mentioned before, our method is the only one that jointly estimates both albedo and shading.
Albedo is a useful property in practice as it allows the usage of our reconstructions in virtual environments with their own light composition. Additionally, as a byproduct of our shading estimation, we can do image compositing~\cite{zanfir2020human, tsai2017deep}, one of the most common photo editing tasks. One example is given in fig.~\ref{fig:sceneedit}. We first computed the illumination $\illum$ from a given target image. We then reconstruct two subjects from studio photographs and use $\illum$ to re-shade them. This allows us to compose a synthesized group picture with matching illumination for all people in the scene.

\subsection{Ablations}
\label{sec:ablations}

We now ablate two main design choices of our method: first, the rendering losses, and second, shading estimation.
In tab.~\ref{tab:numerical_eval}, we report metrics for our method trained without rendering losses (w/o rendering) and without shading estimation (w/o shading).
Furthermore, in fig.~\ref{fig:ablation_quali} we show visual examples of results produced by our model variant trained without rendering losses.

While only using 3D sparse supervision produces accurate geometry, the albedo estimation quality is, however, significantly decreased. As evident in fig.~\ref{fig:ablation_quali} and also numerically in tab.~\ref{tab:numerical_eval}, the estimated albedo contains unnatural color gradient effects.
We hypothesize that due to the sparse supervision, where individual points are projected into the feature map, the feature extractor network does not learn to understand structural scene semantics.
Here our patch-based rendering losses help, as they provide gradients for neighboring pixels. Moreover, our rendering losses could better connect the zero-level-set of the signed distance function with the color field, as they supervise the color at the current zero-level-set and not at the expected surface location.
We plan to structurally investigate these observations, and leave these for future work.

Estimating the shading jointly with the 3D surface and albedo does not impair the reconstruction accuracy.
On the contrary, as evident in tab.~\ref{tab:numerical_eval}, this helps improve albedo reconstruction.
This is in line with our hypothesis that shading estimation helps the networks to better decouple shading effects from albedo. 
Finally, shading estimating makes our method a holistic reconstruction pipeline.

%% file: sections/conclusion.tex
\begin{figure}
    \centering
    \includegraphics[width=1.0\linewidth]{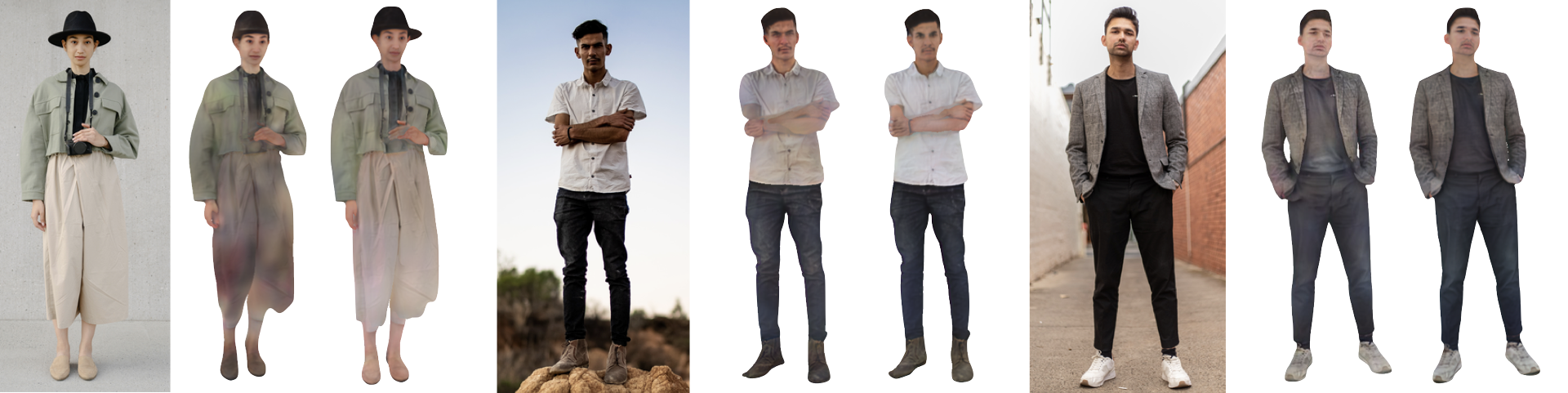}
    \caption{Loss ablation: The usage of our rendering losses (right) significantly improves albedo estimation. Note the unnatural color gradients when using sparse 3D supervision only (left).}
    \label{fig:ablation_quali}
\end{figure}

\begin{figure}
    \centering
    \includegraphics[width=1.0\linewidth]{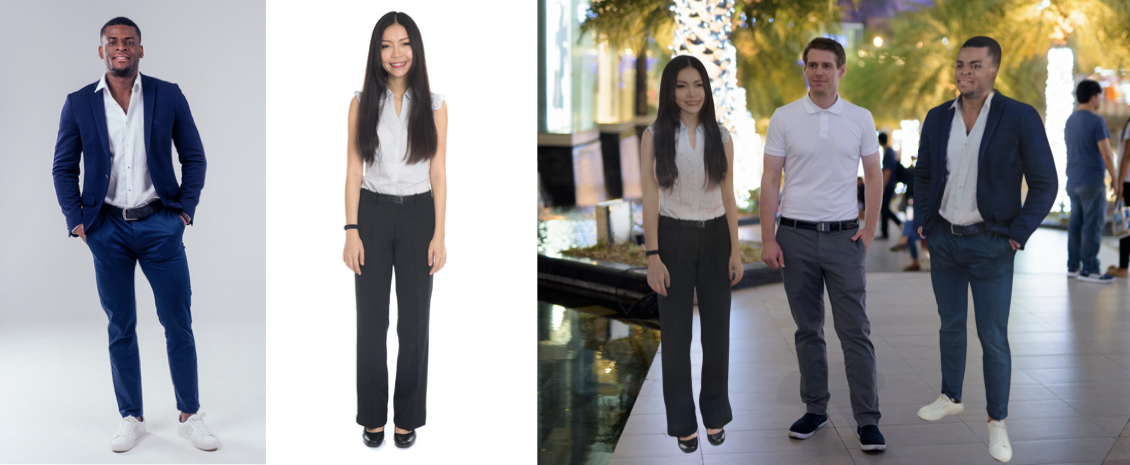}
    \caption{We can apply the estimated illumination from one image to another, which allows us to create the group picture (right) by inserting the reconstructions of the subjects (left) with matching shaded surface.}
    \label{fig:sceneedit}
    \vspace{-1mm}
\end{figure}

\begin{figure}
    \centering
    \includegraphics[width=1.0\linewidth]{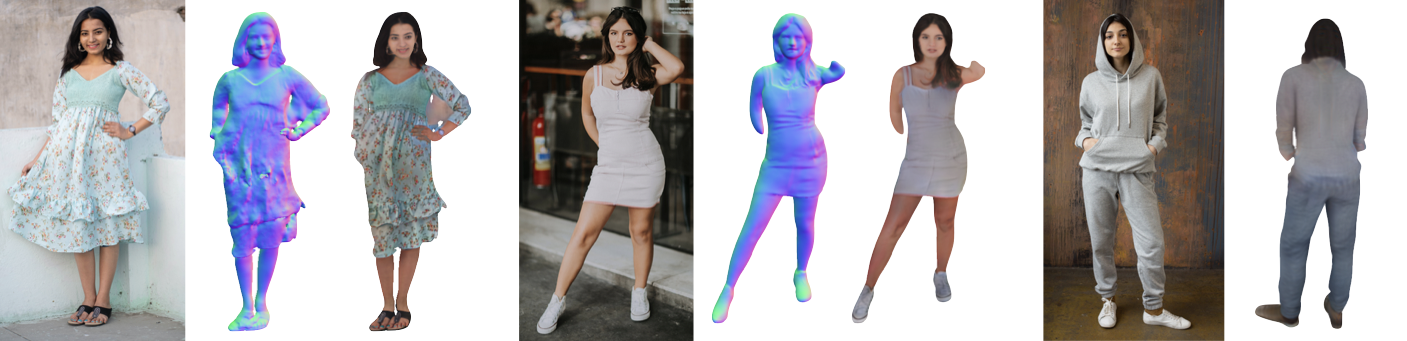}
    \caption{Failure cases. Wide clothing is under-represented in our dataset and this can be addressed with more diverse training. Complex poses can lead to missing body parts. The back-side sometimes mismatches the front (subject is wearing a hood).}
    \label{fig:fail}
    \vspace{-2mm}
\end{figure}

\paragraph{Limitations.}
The limitations of our method are sometimes apparent when the clothing or pose of the person in the input image deviates too much from our dataset distribution, see fig.~\ref{fig:fail}. Loose, oversized, and non-Western clothing items are not well  covered by our training set.
The backside of the person sometimes does not semantically match the front side. A larger, more geographic and culturally diverse dataset would alleviate these problems, as our method does not make any assumptions about clothing style or pose.

\paragraph{Application Use Cases and Model Diversity.} The construction of our
model is motivated by the breadth of transformative, immersive 3D applications, that would become possible, including clothing virtual apparel try-on, immersive visualisation of photographs, personal
AR and VR for improved communication, special effects,
human-computer interaction or gaming, among others. Our models are trained with a diverse and fair distribution, and as the size of this set increases, we expect good practical performance.

\noindent{\bf Conclusions.} We have presented a method to reconstruct the three-dimensional (3D) geometry of a human wearing clothing given a single photograph of that person.
Our method is the first one to compute the 3D geometry, surface albedo, and shading, from a single image, jointly, as prediction of a model trained end-to-end.
Our method works well for a wide variation of outfits and for diverse body shapes and skin tones, and reconstructions capture most of the detail present in the input image.
We have shown that while sparse 3D supervision works well for constraining the geometry, rendering losses are  essential in order to reconstruct perceptually accurate surface color. 
In the future, we would like to further explore weakly supervised differentiable rendering techniques,
as they would support, long-term, the construction of larger and more inclusive models, based on diverse image datasets of people, where accurate 3D surface ground truth is unlikely to be available.

%% file: sections/suppl.tex
In this supplementary material, we detail our implementation by listing the values of all hyper-parameters.
Further, we report inference times, demonstrate how we can repose our reconstructions, conduct further comparisons, and show additional results.

\section{Implementation Details}

In this section, we detail our used hyper-parameters and provide timings for mesh reconstruction via Marching Cubes~\cite{lewiner2003efficient}.

\subsection{Hyper-parameters}

When training the network, we minimize a weighted combination of all defined losses:
\begin{align}
    \loss = \loss_g + \lossweight_e \loss_e + \lossweight_l \loss_l + \loss_{a} + \lossweight_r \loss_r  \nonumber \\ 
    + \lossweight_c \loss_c + \lossweight_s \loss_s + \lossweight_{\text{VGG}} \loss_{\text{VGG}}.
\end{align}
Further, we have defined the weights $\lossweight_{g_1}$, $\lossweight_{g_2}$, $\lossweight_{a_1}$, and $\lossweight_{a_2}$ inside the definitions of $\loss_g$ and $\loss_{a}$.
During all experiments, we have used the following empirically determined configuration: \\
$\lossweight_e=0.1$, $\lossweight_l=0.2$, $\lossweight_r=1.0$, $\lossweight_c=1.0$, $\lossweight_s=50.0$,
$\lossweight_{\text{VGG}} = 1.0$, $\lossweight_{g_2}=1.0$,  $\lossweight_{a_1}=0.5$,  $\lossweight_{a_2}=0.3$ \\
Additionally we found it beneficial to linearly increase the surface loss weight $\lossweight_{g_1}$ from $1.0$ to $15.0$ over the duration of 100k interactions.

\subsection{Inference timings}
To create a mesh we run Marching Cubes over the distance field defined by $\mlp$.
We first approximate the
bounding box of the surface by probing at coarse resolution and use Octree sampling to progressively increase the
resolution as we get closer to the surface.
This allows us to extract meshes with high resolution without large computational overhead.
We query $\mlp$ in batches of $64^3$ samples up to the desired resolution.
The reconstruction of a mesh in a $256^3$ grid takes on average $1.21$s using a single NVIDIA Tesla V100.
Reconstructing a very dense mesh in a $512^3$ grid takes on average $5.72$s.
Hereby, a single batch of $64^3$ samples takes $142.1$ms.
In both cases, we query the features once which takes $243$ms.
In practise, we also query $\mlp$ a second time for color at the computed vertex positions which takes $56.5$ms for meshes in $256^3$ and $223.3$ms for $512^3$, respectively.
Meshes computed in $256^3$ and $512^3$ grids contain about 100k and 400k vertices, respectively.
Note that we can create meshes in arbitrary resolutions and our reconstructions can be rendered through sphere tracing without the need to generate an explicit mesh.

\begin{figure*}
    \centering
    \captionsetup{width=.8\linewidth}
    \includegraphics[width=0.73\linewidth]{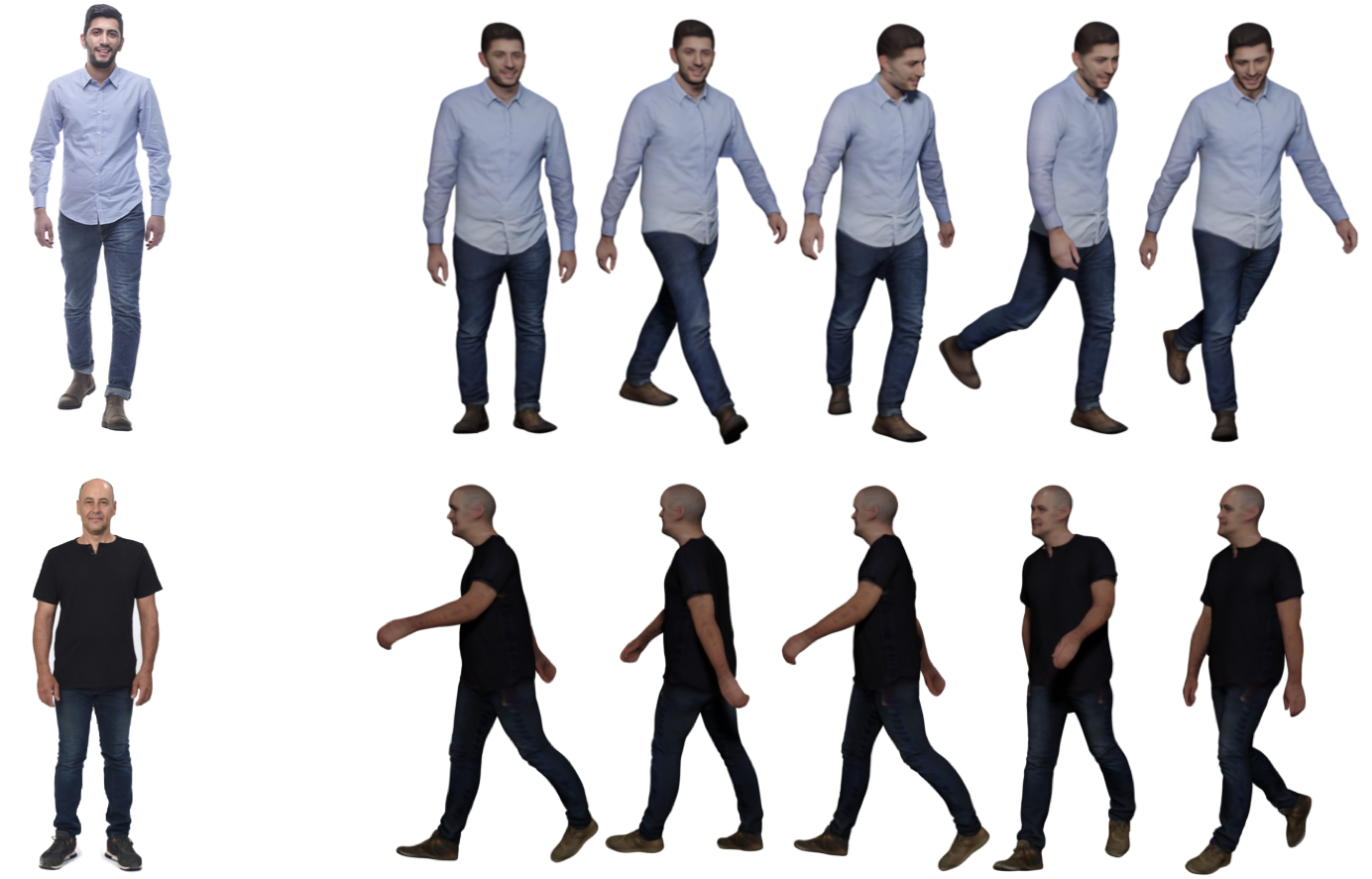}
    \caption{Examples of reconstructions rigged and animated in a post processing step. We show the input image (left) and re-posed reconstructions (right). The reconstructions are rendered under a novel illumination.}
    \label{fig:anim}
	\vspace{5mm}

    \includegraphics[width=0.8\linewidth]{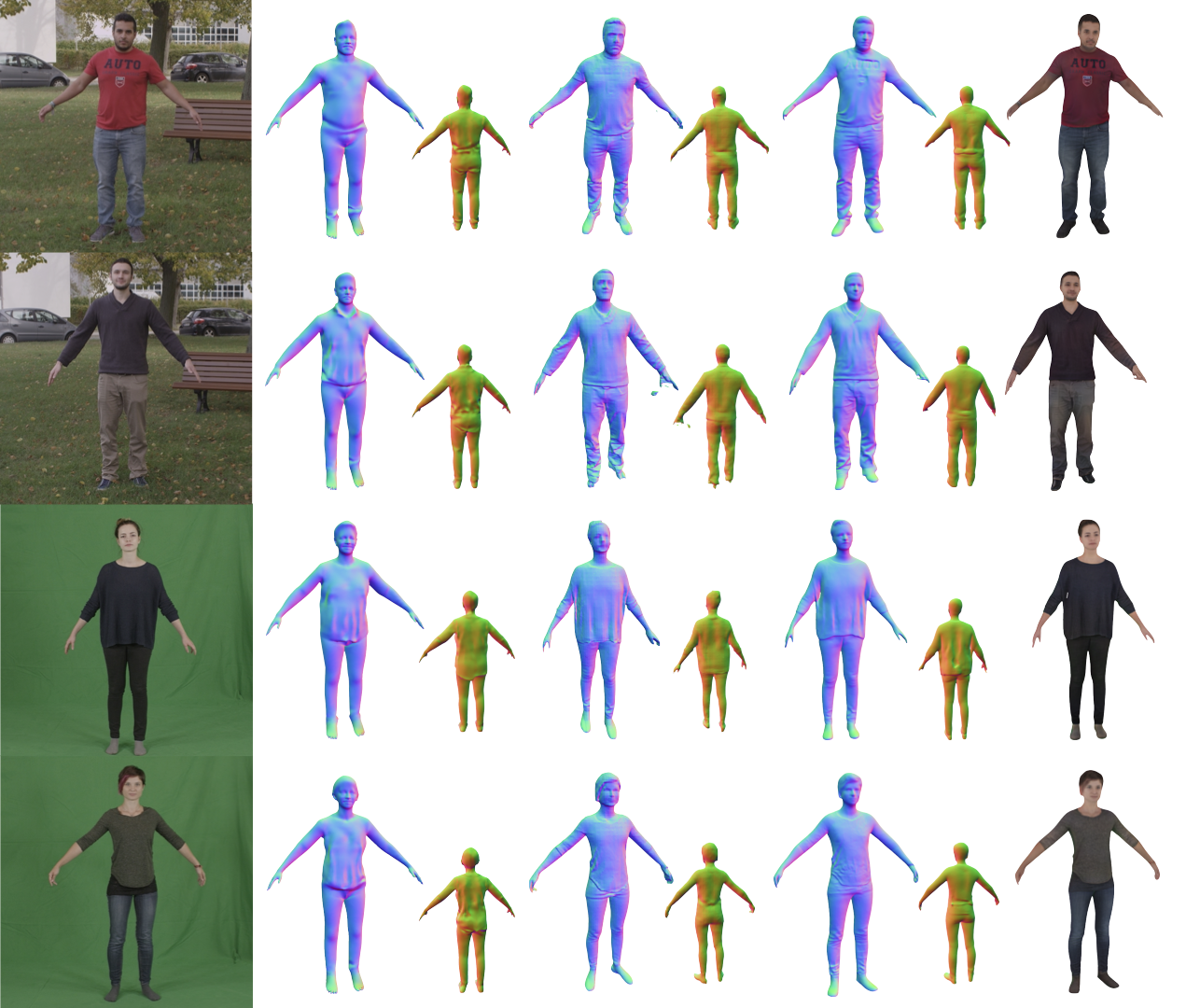}
    \caption{Qualitative comparison on the PeopleSnapshot dataset \cite{peoplesnapshot}. From left to right: Input image, geometry produced by Tex2Shape \cite{alldieck2019tex2shape}, PIFuHD \cite{saito2020pifuhd}, and PHORHUM (ours). We additionally show albedo reconstructions for our method.}
    \label{fig:suppl_peoplesnapshot}
\end{figure*}

\section{Additional Results}

In the sequel, we show additional results and comparisons. First, we demonstrate how we can automatically rig our reconstructions using a statistical body model. Then we conduct further comparisons on the PeopleSnapshot Dataset \cite{peoplesnapshot}. Finally, we show additional qualitative results.

\subsection{Animating Reconstructions}
In fig.~\ref{fig:anim}, we show examples of rigged and animated meshes created using our method.
For rigging, we fit the statistical body model GHUM~\cite{xu2020ghum} to the meshes.
To this end, we first triangulate joint detections produced by an off-the-shelf 2D human keypoint detector on renderings of the meshes.
We then fit GHUM to the triangulated joints and the mesh surface using ICP.
Finally, we transfer the joints and blend weights from GHUM to our meshes.
We can now animate our reconstructions using Mocap data or by sampling GHUM's latent pose space.
By fist reconstructing a static shape that we then rig in a secondary step, we avoid reconstruction errors of methods aiming for animation ready reconstruction in a single step \cite{huang2020arch, he2021arch++}.

\subsection{Comparisons on the PeopleSnapshot Dataset}
We use the public PeopleSnapshot dataset \cite{alldieck2018video,peoplesnapshot} for further comparisons.
The PeopleSnapshot dataset contains of people rotating in front of the camera while holding an A-pose.
The dataset is openly available for research purposes.
For this comparison we use only the first frame of each video.
We compare once more with PIFuHD \cite{saito2020pifuhd} and additionally compare with the model-based approach Tex2Shape \cite{alldieck2019tex2shape}. Tex2Shape does not estimate the pose of the observed subject but only its shape.
The shape is represented as displacements to the surface of the SMPL body model \cite{smpl2015loper}.
In fig.~\ref{fig:suppl_peoplesnapshot} we show the results of both methods side-by-side with our method.
Also in this comparison our method produces the most realistic results and additionally also reconstructs the surface color.

\subsection{Qualitative Results}
\begin{figure*}
    \centering
    \captionsetup{width=.85\linewidth}
    \includegraphics[width=0.85\linewidth]{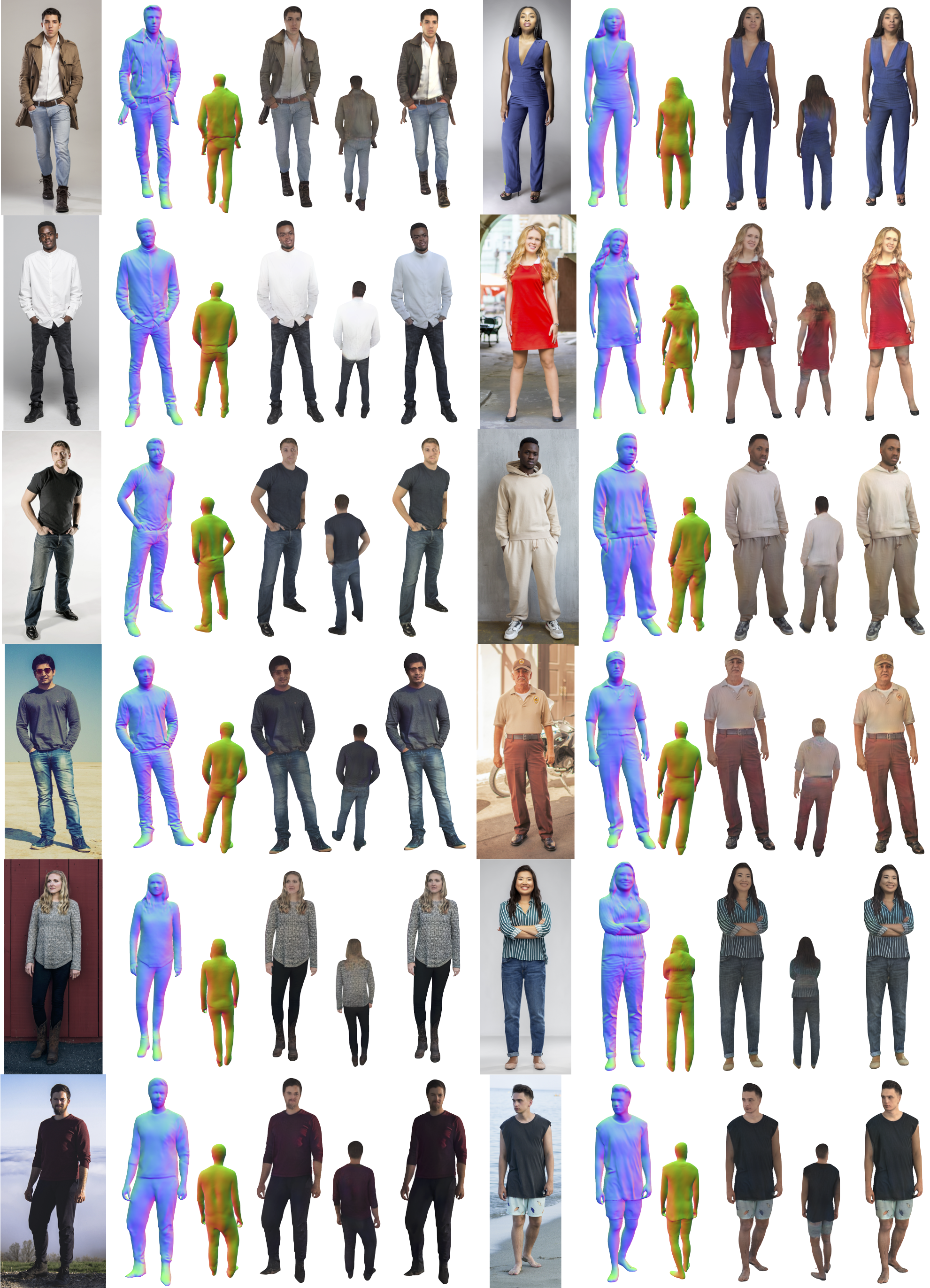}
    \caption{Qualitative results on real images featuring various outfits, backgrounds, and illumination conditions. From left to right: Input image, 3D geometry (front and back), albedo reconstruction (front and back), and shaded surface.}
    \label{fig:suppl_quali_bg}
    \vspace{5mm}
\end{figure*}

\begin{figure*}
    \centering
    \captionsetup{width=.85\linewidth}
    \includegraphics[width=0.85\linewidth]{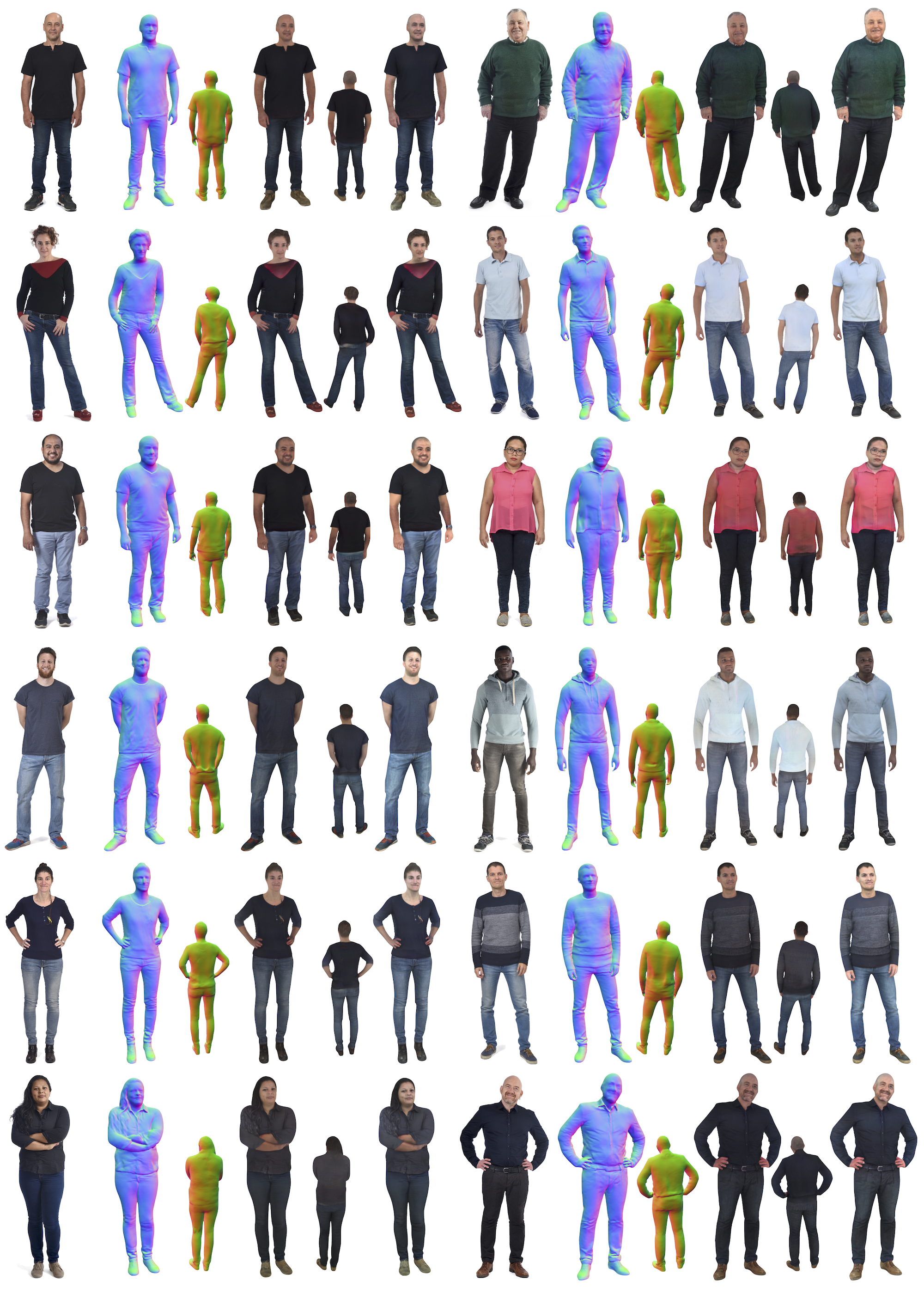}
    \caption{Despite never being trained on matted images, our method performs extremely well on images with white background. From left to right: Input image, 3D geometry (front and back), albedo reconstruction (front and back), and shaded surface.}
    \label{fig:suppl_quali_nobg}
\end{figure*}

We show further qualitative results in fig.~\ref{fig:suppl_quali_bg}.
Our methods performs well on a wide range of subjects, outfits, backgrounds, and illumination conditions.
Further, despite never being trained on this type of data, our method performs extremely well on image of people with solid white background.
In fig.~\ref{fig:suppl_quali_nobg} we show a number of examples.
This essentially means, matting the image can be performed as a pre-processing step to boost the performance of our method in cases where the model has problems identifying foreground regions.

